\documentclass{article}

\usepackage{PRIMEarxiv}

\usepackage[utf8]{inputenc} 
\usepackage[T1]{fontenc}    
\usepackage{hyperref}       
\hypersetup{
    colorlinks=true,
    linkcolor=blue,
    filecolor=blue,      
    urlcolor=blue,
    citecolor=blue,
    pdfpagemode=FullScreen,
    }
\usepackage{url}            
\usepackage{booktabs}       
\usepackage{amsfonts}       
\usepackage{nicefrac}       
\usepackage{microtype}      
\usepackage{lipsum}
\usepackage{fancyhdr}       
\usepackage{graphicx}       
\graphicspath{{media/}}     

\usepackage{natbib}
\usepackage{amssymb}
\usepackage{amsthm}
\usepackage{mathtools}
\usepackage{comment}
\DeclareMathOperator{\bE}{\mathbb{E}}

\usepackage{xcolor}
\usepackage[ruled,vlined]{algorithm2e}

\usepackage{xcolor}
\usepackage[ruled,vlined]{algorithm2e}
\usepackage{graphicx}

\newcommand{\watchtime}{\texttt{WatchTime}}
\newcommand{\share}{\texttt{Share}}

\newcommand{\download}{\texttt{Download}}
\newcommand{\cmt}{\texttt{Comment}}

\usepackage[utf8]{inputenc} 
\usepackage[T1]{fontenc}    
\usepackage{hyperref}       
\usepackage{url}            
\usepackage{booktabs}       
\usepackage{amsfonts}       
\usepackage{nicefrac}       
\usepackage{microtype}      
\usepackage{xcolor}         
\usepackage{amssymb}
\usepackage{amsthm}
\usepackage{mathtools}
\usepackage{comment}

\title{Constrained Reinforcement Learning for Short Video Recommendation
}

\author{
  Qingpeng Cai, Ruohan Zhan,  Chi Zhang, Jie Zheng, Guangwei Ding, \\ \textbf{Pinghua Gong, Dong Zheng, Peng Jiang} \\
  Kuaishou Technology \\
  Beijing, China \\
  \texttt{\{caiqingpeng,zhanruohan,zhangchi08,zhengjie,dingguangwei03},\\ \texttt{gongpinghua,zhengdong,jiangpeng\}@kuaishou.com} \\
}

\begin{document}
\maketitle



%


\begin{abstract}
  The wide popularity of short videos on social media poses new opportunities and challenges to optimize recommender systems on the video-sharing platforms. Users provide complex and multi-faceted responses towards recommendations, including watch time and various types of interactions with videos. As a result, established recommendation algorithms that concern a single objective are not adequate to meet this new demand of optimizing comprehensive user experiences. In this paper, we formulate the problem of short video recommendation as a \emph{constrained} Markov Decision Process (MDP), where platforms want to optimize the main goal of user watch time in long term, with the constraint of  accommodating  the auxiliary responses of user interactions such as sharing/downloading videos. 
    
    To solve the constrained MDP, we propose a two-stage  reinforcement learning approach based on actor-critic framework. At stage one, we learn individual policies to optimize each auxiliary response. At stage two, we learn a policy to (i) optimize the main response and (ii) stay close to policies learned at the first stage, which effectively guarantees the performance of this main policy on the auxiliaries. Through extensive simulations, we demonstrate effectiveness of our approach over alternatives in both optimizing the main goal as well as balancing the others. We further show the advantage of our approach in live experiments of short video recommendations, where it significantly outperforms other baselines  in terms of watch time and interactions from video views. Our approach has been fully launched in the production system to optimize user experiences on the platform. 
\end{abstract}

\section{Introduction}
The surging popularity of short videos has been changing the status quo of social media.
As of 2021, the monthly active users on TikTok have reached one billion worldwide \cite{TikTok}. Such prevalence of short video consumption has brought in huge business opportunities for organizations. As a result, there has been an increasing interest in optimizing recommendation strategies for short video platforms, where  user feedback is multifaceted. Potential responses from a user after consuming a video include \watchtime~(the time spent on watching the video), 
and several types of interactions:
\share~(sharing this video with his/her friends), \download~(downloading the video), \cmt~(providing comments on the video), etc. 
Thereby, established recommender systems that exclusively optimize a single objective (such as gross merchandise volume for e-commence platforms \cite{pi2020search}) is no longer sufficient---the applied systems  should take all aspects of responses into consideration to optimize user experiences. 


In this paper, we present our solution in the context of constrained optimization. As opposed to Pareto optimality that is often applied to study multi-objective strategies \cite{sener2018multi,chen2021reinforcement}, preferences on different objectives are often pre-specified in real applications.  
Notably, one main goal for short video platforms is to increase the watch time, 
which is observed from each video view and widely concerns all  users. Besides, watch time reflects user attention, which is  the scarce resource that companies compete for. 
Conversely, other responses such as \share/\cmt~are not mutually exclusive among platforms and thus could be sacrificed mildly.
On the other hand, platforms have been focusing on optimizing user long-term engagement, which directly drives daily active users (DAU) and thereby the revenue growth. Recently, a growing literature has focused on applying reinforcement learning (RL) to recommender systems, due to   its ability to improve cumulative reward \cite{nemati2016optimal,zhao2017deep,  zhao2018recommendations,chen2018stabilizing, zou2019reinforcement,liu2019deep, chen2019large, xian2019reinforcement, ma2020off, afsar2021reinforcement, ge2021towards}. In particular, watch time, as the dense response, can be effectively cumulatively maximized to increase user spent time across multiple requests with RL approaches \cite{chen2019top}. 
Thereby, we propose to learn an RL-based agent that optimizes the \emph{main} goal (\watchtime), with the constraint of compensating other \emph{auxiliary} responses (\share, \download, and \cmt) with reasonable levels.

The problem of this constrained policy learning is much more challenging as compared to its unconstrained counterpart. A natural idea would be learning a value-based or policy-based model that maximizes the Lagrangian with pre-specified multipliers. However, such method is often difficult to be realized in practice via standard RL methods for the following two reasons.

First, it is not sufficient to use a single policy evaluation model to estimate 
the Lagrangian dual objective. As discussed, the agent may receive different types of responses from the user. A straightforward approach is to combine them into a single weighted sum using pre-specified multipliers, and learn a value-based model such as Q-learning \cite{mnih2013playing} to optimize it, as proposed in \cite{stamenkovic2021choosing}. 
Such response combination is not adequate, particularly for responses with their own discount factors---the formulation of temporal difference error in value-based models  only allows for a single discount value. In scenarios where one discount factor suffices, it can still be difficult for a single value model to evaluate the policy accurately, especially when different responses are observed at various frequencies, as typical for short video recommendations. The \watchtime~response is dense and observed from each video view, while the interaction-signal such as \share/\cmt~is much more sparse and may not be provided within dozens of views. Signal from the sparse responses will be weakened by the dense responses when naively summing them up together.  To address this multi-response evaluation difficulty,  we  separately evaluate  each response via its own value model, which allows for response-specific discount factors and mitigates the interference on evaluation from one response on another, similar to the procedure conducted in \cite{chen2021reinforcement,tajmajer2018modular,hessel2019multi}. As an example, we evaluate the behavior policy on a popular short video platform using data collected real time and find that  such separate evaluation improves learning on   \watchtime~and interaction-signal by $0.191\%$ and  $0.143\%$ respectively{\footnote{In real applications for video recommendations, an improvement around $0.1\%$ on value estimation is already significant to be reflected in production performance.}}; Appendix A elaborates the experimental detail. 

Second, it is hard for a single policy to balance both dense responses and sparse responses.  Learning a sparse response itself is well-known to be problematic---it may take the agent undesirably long to learn something meaningful \cite{florensa2017reverse,riedmiller2018learning}.  Coexistence of both dense and sparse responses exacerbates the learning difficulty. In most time, the agent only learns to optimize the policy in the direction of optimizing dense responses, which may negatively affect its learning for  sparse responses.   On account of this, we propose to firstly learn a policy to optimize each auxiliary response and then ``softly'' regularize the policy of the main response to be in the neighborhood of  others. We demonstrate empirically that our approach can better balance different responses in both simulated data and live experiments. 

Together, we summarize our contributions as below:
\begin{itemize}
    \item \textbf{Constrained Optimization in Short Video Recommendations}: We formalize the problem of constrained policy learning in short video recommendations, where different aspects of responses may be observed at various frequencies, and the agent learns to optimize one with the constraint of balancing others.
    \item \textbf{Multi-Critic Policy Estimation}: To better evaluate policy on multiple responses that may differ in discount factors and observation frequencies, we propose to separately learn a value model to evaluate each response.
    \item \textbf{Two-Stage Actor-Critic Learning}: We propose a two-stage actor-critic framework which firstly learns a policy to optimize each auxiliary response and secondly regularizes the policy of the main response to be not far from others, which we demonstrate to be a more effective way in constrained optimization as compared with other alternatives. 
    \item \textbf{Gains in Live Experiments}: We demonstrate the effectiveness of our approach in live experiments, showing the ability of our approach in optimizing  the main response of \watchtime~as well as balancing other interaction ones. 
\end{itemize}

\section{Related Work}

\paragraph{Constrained Reinforcement Learning} Our work is also closely related to the literature of  constrained reinforcement learning, where the sequential decision making problem is formulated into a constrained Markov Decision Process \cite{sutton2018reinforcement}, and the policy learning procedure is expected to respect the constraints. There are mainly two categories of constraints:  cumulative ones (sum of a given signal should be limited into certain region) and instantaneous ones (constraints should be 
satisfied at each step) \cite{liu2021policy,perkins2002lyapunov,garcia2015comprehensive}. To deal with cumulative constraints, there is a large body of literature focusing on Lagrangian relaxation \cite{chow2017risk,chow2019lyapunov,tessler2018reward,dalal2018safe}. As an example, \cite{tessler2018reward} propose to update the policy and the Lagrangian multiplier alternatively and prove the convergence of their algorithm to a fix point. This approach however does not deal with the difficulty of policy learning on rewards with different observation frequencies and thus is difficult to achieve a good balance among multiple responses. In contrast, for each cumulative reward, we learn a policy to maximize it specifically, then we  ``softly'' regularize the main policy to be in the neighborhood of others. We show empirically that this is a more effective way for constrained policy learning when dealing with both sparse and dense rewards. Different from \cite{nair2020awac} that studies in offline RL and regularizes the learned policy to be in the neighborhood of one behavior policy, we softly restrict the policy within other policies maximizing other auxiliary responses and we do not limit to offline settings.

\paragraph{Multi-objective Optimization} We also discuss a relevant line on multi-objective optimization. To trade off different objectives, methods in this field can be broadly categorized into two classes: the Pareto optimization and the joint optimization with pre-specified weights. The goal of Pareto optimization is to find a solution such that no other solutions can concurrently improve all objectives, named as \emph{Pareto optimality} \cite{nguyen2020multi,sener2018multi,chen2021reinforcement,ge2022toward}. However, a Pareto optimal solution may not prioritize the objective that is most valued in  applications. 
The second method combines different objectives  together into a single one via pre-specify the weights \cite{white1980solution,mossalam2016multi}. However, it is difficult to quantify these weights that can accurately reflect preferences in real applications \cite{tessler2018reward}.

\section{Preliminaries}
\subsection{Constrained Markov Decision Process}
\label{sec:cmdp}
We start by formulating the problem of short video recommendation on mobile app services. When a user $u$ opens the app, a new \emph{session} starts. A session consists of multiple \emph{requests}. At each request $t$ when the user slides down the app, the recommender system (agent) takes an \emph{action} $a_t$ that recommends the user a video based on the user current \emph{state}, characterized by his/her demographics, historical interactions, etc. Then the user provides \emph{multi-faceted} responses (such as  \watchtime, \share, \download, \cmt) on the shown video, which are received by the agent as vector-valued \emph{reward} signal and used for future planning; let $m$ be the number of types of responses. The goal of the recommender system is to optimize cumulative reward of the main response (\emph{e.g.}, \watchtime), with the constraint of not sacrificing others much. 

We model the above procedure as a Constrained Markov Decision Process(CMDP) \cite{sutton2018reinforcement} $(S, A, P, R, C, \rho_0, \Gamma)$, 
where $S$ is the state space of user current representation $s_t$, 
$A$ is the action space (and each action $a_t$ corresponds to a recommended  video for one request), 
$P:S\times A \rightarrow \Delta(S)$ captures  the state transition, 
$R: S\times A \rightarrow  \mathbb{R}^m$ defines the vector-valued reward function that yields $m$ different rewards $r(s_t, a_t)=\big(r_{1}(s_t, a_t), \dots, r_{m}(s_t, a_t)\big)$, 
$\rho_0$ is the initial state distribution, 
$\Gamma=(\gamma_{1}, \dots, \gamma_{m}) \in (0,1)^m$ denotes the  vector of discount factor for reward of each response, and $C$ specifies the constraints on the auxiliary responses. 

Define the  vector-valued discounted cumulative reward $R_t$  as ${{R}}_t = \sum_{t'=t}^T \Gamma^{t'-t} \cdot { r}(s_{t'}, a_{t'})$,
where $T$ is the session length (i.e., the number of requests), $\Gamma^{b}=\big(\gamma_{1}^b, \dots, \gamma_{m}^s\big)$, and $\mathbf{x}\cdot \mathbf{y}$ denotes the pointwise product. Let $V^{\pi}(s)=\big(V_{1}^\pi(s),\dots,  V_{m}^\pi(s)\big)$ be the state value $E_{\pi}[R_t|s_t=s]$ under actions sampled in accordance with policy $\pi$ and $Q(s, a)=\big(Q_{1}^\pi(s,a),\dots,  Q_{m}^\pi(s,a)\big)$
be its state-action value $E_{\pi}[R_t|s_t=s, a_t=a]$. Denote $\rho_\pi$ as the state distribution induced by policy $\pi$.
Without loss of generality, we set the first response as our main response. The goal is to learn a recommendation policy $\pi(\cdot|s)$ over the action space to solve the following optimization problem: 
\begin{equation}
\label{eq:prob}
\begin{split}
    \max_\pi \quad &  E_{\rho_\pi}\big[V^{\pi}_{1}(s)\big] \\
    \mbox{s.t.}   \quad &  E_{\rho_\pi}\big[V^{\pi}_{i}(s)\big]  \geq C_{i}, \quad i=2,\dots,m
\end{split}
\end{equation}
where $ C_{i}$ is constraint on the \emph{auxiliary} response $i$. 

\section{Two-Stage Constrained Actor Critic}
In this section, we propose our two-stage constrained policy learning based on actor-critic framework, addressing the learning challenges in the context of dense and sparse rewards:
\begin{description}
    \item[Stage One] For each auxiliary response, we learn a policy to optimize its cumulative reward.  
    \item[Stage Two] For the main response, we learn a policy to  optimize its cumulative reward, while limiting it to be close to other policies that are learned to optimize the auxiliary. 
\end{description}
We first elaborate our framework in the settings of online learning with stochastic policies (such as A2C and A3C \cite{williams1992simple,mnih2016asynchronous}) in Sections \ref{sec:stage-one} and \ref{sec:stage-two}(the procedure is summarized in Appendix C). We then discuss its extensions to deterministic policies (such as DDPG and TD3 \cite{lillicrap2015continuous,fujimoto2018addressing}).
For offline setting, please refer to 
Appendix D.

\subsection{Stage One: Policy Learning for Auxiliary Responses}
\label{sec:stage-one}
At this stage, we learn  policies to optimize the cumulative reward of each auxiliary response separately. For completeness, we write out our procedure following the standard advantage actor critic approach \cite{williams1992simple}. Considering response $i$, let the learned actor and the critic be parameterized by $\pi_{\theta_i}$ and $V_{\phi_i}$ respectively. At iteration $k$, we observe sample $(s,a,s')$ collected by $\pi_{\theta_i^{(k)}}$, \emph{i.e.}, $s\sim \rho_{\pi_{\theta_i^{(k)}}}, a\sim \pi_{\theta_i^{(k)}}(\cdot|s)$ and $s'\sim P(\cdot|s,a)$. We update the critic to minimize the Bellman equation:
\begin{equation}
\label{eq:nonmajor-critic}
    \phi_{i}^{(k+1)} \leftarrow \arg\min_{\phi}
    E_{\pi_{\theta_i^{(k)}}}\Big[
    \big(r_{i}(s, a) + \gamma_{i}V_{\phi_i^{(k)}}(s') -V_{\phi}(s)  \big)^2
    \Big].
\end{equation}
We update the actor to maximize the advantage:
\begin{equation}
\label{eq:nonmajor-actor}
\begin{split}
    &\theta_{i}^{(k+1)} \leftarrow \arg\max_{\theta}  E_{\pi_{\theta_i^{(k)}}}\Big[ A_{i}^{(k)}\log\big(\pi_\theta(a|s)\big)\Big]\\
    \mbox{where}\quad & A_{i}^{(k)} = r_{i}(s, a) +\gamma_{i} V_{\phi_i^{(k)}}(s') -V_{\phi_i^{(k)}}(s).
\end{split}
\end{equation}

\subsection{Stage Two: Constrained Optimization of the Main Response}
\label{sec:stage-two}
After pretraining the policies $\pi_{\theta_2}, \dots, \pi_{\theta_m}$ that optimize the auxiliary responses, we now move onto the second stage of learning the policy to optimize the main response. We propose a new constrained advantage actor critic approach.
Let the actor and the critic be $\pi_{\theta_1}$ and $V_{\phi_1}$ respectively. At iteration $k$, we similarly update the critic to minimize the Bellman equation:
\begin{equation}
\label{eq:major-critic}
     \phi_{1}^{(k+1)} \leftarrow \arg\min_{\phi}
    E_{\pi_{\theta_1^{(k)}}}\Big[
    \big(r_{1}(s, a) + \gamma_{1}V_{\phi_1^{(k)}}(s') -V_\phi(s)  \big)^2
    \Big].
\end{equation}

The principle of updating the actor is two-fold: (i) maximizing the advantage; (ii) restricting the policy to the domain that is not far from other policies. The optimization is formalized below:
\begin{equation}
\label{eq:awac}
    \begin{split}
         \max_{\pi}\quad  & E_{\pi}[ A_{1}^{(k)}]\\
         \mbox{s.t.} \quad & D_{KL}(\pi|| \pi_{\theta_i})\leq \epsilon_i,\quad i=2,\dots,m,\\
    \mbox{where}\quad & A_{1}^{(k)} = r_{1}(s, a) +\gamma_{1} V_{\phi_1^{(k)}}(s') - V_{\phi_1^{(k)}}(s).
    \end{split}
\end{equation}
Equation \eqref{eq:awac} has the closed form solution 
\begin{equation}
  \pi^*(a|s)  \propto \prod_{i=2}^m \big(\pi_{\theta_i}(a|s)\big)^{\frac{\lambda_i}{\sum_{j=2}^m \lambda_j}}\exp\bigg(\frac{A_{1}^{(k)}}{\sum_{j=2}^m \lambda_j} \bigg),
\end{equation}
where $\lambda_i$ with $i=2,\dots,m$ are Lagrangian multipliers for constraints in \eqref{eq:awac},
and the value of $\lambda_i$ controls the degree of constraint. 

Given data collected by $\pi_{\theta_1^{(k)}}$, we learn the policy $\pi_{\theta_1}$ by minimizing its KL divergence from the optimal policy $\pi^*$:

\begin{equation}
\label{eq:major-actor}
\begin{split}
     \theta_1^{(k+1)} \leftarrow& \arg\min_{\theta}E_{\pi_{\theta_1^{(k)}}}[D_{KL}(\pi^*(\cdot|s)||\pi_\theta(\cdot|s))]\\
     = &\arg\max_{\theta}E_{\pi_{\theta_1^{(k)}}}\Big[ 
     \prod_{i=2}^m \Big(\frac{\pi_{\theta_i}(a|s)}{\pi_{\theta_1^{(k)}}(a|s)}\Big)^{\frac{\lambda_i}{\sum_{j=2}^m \lambda_j}}\exp\bigg(\frac{A_{1}^{(k)}}{\sum_{j=2}^m \lambda_j} \bigg)
    \log \pi_\theta(a|s) \Big].
\end{split}
\end{equation}

Appendix B contains the derivation details. We here provide some intuition behind actor updating in \eqref{eq:major-actor}. The ratio $\frac{\pi_{\theta_i}(a|s)}{\pi_{\theta_1^{(k)}}(a|s)}$ suggests that the updating direction of policy $\pi_{\theta_1}$ will be favored when it's aligned with the constraint policies $\pi_{\theta_i}$, which effectively regularizes  the learned policy $\pi_{\theta_1}$ to be in the neighborhood of other policies $\pi_{\theta_i}$. Small Lagrangian multipliers $\lambda_i$ indicate weaker constraints, and when $\lambda_i=0$, we allow the learned policy $\pi_{\theta_1}$ to be irrelevant of the constraint policy $\pi_{\theta_i}$.

\textbf{Deterministic Policies}
\label{sec:ddpg}
We now  shed light on  adaptation of our framework to deterministic policies  such as deep deterministic policy gradient (DDPG) algorithm \cite{lillicrap2015continuous}, inspired by the updating rule for the actor of constrained policy discussed in \eqref{eq:major-actor}. Similarly, at stage one, for each auxiliary response $i$, we learn the actor $\pi_{\theta_i}(s)$ and critic $Q_{\phi_i}(s,a)$ via DDPG algorithm respectively.  At stage two, for the main response, we learn critic $Q_{\phi_1}(s,a)$ via temporal learning; and for   actor $\pi_{\theta_1}(s)$,
the updating rule follows the form of
\begin{equation}
    \max_{\theta_1}\quad \prod_{i=2}^m \bigg(\frac{ h(a,\pi_{\theta_i}(s))}{h(a,\pi_{\theta_i}(s))}\bigg)^{\frac{\lambda_1}{\sum_{j=2}^m\lambda_j}} f\bigg(\frac{Q_{\phi_1}(s, \pi(s))}{\sum_{j=2}^m \lambda_j}\bigg),
    \label{eq:ddpg}
\end{equation}
where $f$ is an increasing function which pushes the gradient of $\pi_{\theta_1}$ towards increasing the policy value; $h(a_1, a_2)$ scores high when two actions $a_1, a_2$ are close to each other and scores low vice versa; $\lambda_i\geq 0$ plays the same role as the constraint Lagrangian multiplier in \eqref{eq:major-actor}---larger $\lambda_i$ denotes stronger constraint. As a demonstration, one can choose $f$ to be the identity function and $h(a_1, a_2)=\exp\big(-\frac{(a_1-a_2)^2}{2}\big)$. Section \ref{sec:offline} showcases how this construction of softly constrained DDPG algorithm effectively  achieves the main goal as well as balancing the auxiliaries.

\section{Offline Experiments}
\label{sec:offline}
In this section, we evaluate our approach on a public dataset via extensive offline learning simulations. We demonstrate the effectiveness of our approach as compared to existing baselines in both achieving the main goal and balancing the auxiliaries.

\subsection{Setup}
\paragraph{Dataset} We consider a  hotel-review dataset named \emph{TripAdvisor}, which is a standard dataset for studying policy optimization in recommender system with multiple responses in \cite{chen2021reinforcement}. 
In this data, customers not only provide an \emph{overall} rating for hotels but also score hotels in multiple aspects including \emph{service}, \emph{business}, \emph{cleanliness}, \emph{check-in}, \emph{value}, \emph{rooms}, and \emph{location} \cite{alam2016joint}.
\footnote{The dataset consists of both the main objective and other responses, which can also be used to evaluate constrained policy optimization in recommender system.} Reviews provided by the same user are concatenated chronologically to form a trajectory; we filter trajectories with length smaller than $20$.
In total, we have $20277$ customers, $150$ hotels, and $257932$ reviews. 

\paragraph{MDP} A trajectory tracks a customer hotel-reviewing history.  For each  review, we have
state $s_t$: customer ID and  the last three reviewed hotel IDs as well as corresponding multi-aspect review scores; action $a_t$: currently reviewed hotel ID; reward $r_t$: a vector of eight scores the customer provided for the reviewed hotel in terms of \emph{service}, \emph{business}, \emph{cleanliness}, \emph{check-in}, \emph{value}, \emph{rooms},  \emph{location}, and \emph{overall rating}; discount factor $\gamma$: 0.99. We set the main goal to be maximizing the cumulative overall rating, and treat others as the auxiliaries. 

\paragraph{Evaluation}  We use the \emph{Normalised Capped Importance Sampling} ((NCIS) approach to evaluate different policies, which is a standard offline evaluation method in literature \cite{swaminathan2015self}. 

\paragraph{Compared algorithms}\label{sec:alg} We compare our approach with a range of recommendation algorithms.
\begin{itemize}
    \item \textbf{BC}: a supervised behavior-cloning policy $\pi_\beta$ to mimic customer reviewing pattern, with input as the user state and output as the reviewed hotel ID. 
    \item \textbf{Wide\&Deep}\cite{cheng2016wide}: a supervised model which utilizes wide and deep layers to balance both memorization and generalization, with  input as the user state, output as the reviewed hotel id, and sample weight as the weighted sum of $8$ scores for this review. 
    \item \textbf{A3C}\cite{mnih2016asynchronous}: an online RL  approach with one actor and one critic, where reward  is the  weighted sum of $8$ scores for a given customer-hotel review.
    \item \textbf{DDPG}\cite{lillicrap2015continuous}: an offline RL approach with one actor and one critic, where reward is the  weighted sum of $8$ scores for a given customer-hotel review. 
    \item \textbf{RCPO}: an offline RL approach that extends the reward-constrained policy optimization of A3C algorithm in \cite{tessler2018reward} to DDPG algorithm. Contrary to the standard DDPG, we learn eight critics for the eight scores and use Lagrangian multipliers to sum them up for the actor optimization.  
    \item \textbf{Pareto}: a recommendation model based on DDPG algorithm to find the Pareto optimal solution for multi-objective optimization.
    \item \textbf{Constrained (Ours)}: our constrained actor critic approach based on DDPG algorithm, where the construction follows the discussion in Section \ref{sec:ddpg}. 
\end{itemize}
 We note that we use DDPG instead of A3C as the base actor critic model to develop constrained policy optimization (RCPO, Pareto, and Ours), by the nature of offline learning. As a comparison, we also present the performance of A3C, which is for online learning and thus is outperformed by DDPG on this dataset---as we shall see shortly. 

\subsection{Overall Performance}

\begin{table*}
    \centering
     \renewcommand\arraystretch{1.1} 
    \begin{tabular}{c|ccc cccc|c}
    \hline
        Algorithm &  BC &  Wide\&Deep & A3C & DDPG & RCPO & Pareto & Constrained  \\
  \hline
  Service	&	$3.38$	& 	$3.41^{*}$	&		$3.37$	&	$3.4$	&	$3.41$	&	$3.36$	& $\mathbf{3.43}$ \\  
  \hline
  Business	&	$-1.86$	& 	$-1.86$	& $-1.78^{*}$	&	$-1.82$	&	$-1.82$	&	$-1.79$	&	-1.82		\\
  \hline
  Cleanliness	&	$3.57$	& 	$3.62^{*}$	&		$3.56$	&	$3.61$	&	$3.62^{*}$	&	3.57	&	$\mathbf{3.64}$		\\
  \hline
  Check-in	&	$-0.73$	& 	$-0.75$	&		$-0.65$	&	$-0.71$	&	$-0.68$	&	$-0.62^{*}$	&	$-0.68$		\\
  \hline
  Value	&	$3.32$	& 	$3.36^{*}$	&		$3.27$	&	$3.34$	&	$3.35$	&	$3.29$	&	$\mathbf{3.37}$		\\
  \hline
  Rooms	&	$2.92$	& 	$2.96$	&	$2.97^{*}$		&	$2.95$	&	$2.97^{*}$	&	$2.93$	&	\textbf{3.00}		\\ \hline
  Location	&	$2.93^{*}$	& 	$2.88$	& 	$2.88$	&	$2.91$	&	$2.87$	&	$2.86$	&	\textbf{2.98}		\\ \hline
  Overall Rating	&	$3.92$	& 	$3.98$		& 	$3.94$	&	$3.97$	&	$\mathbf{3.99}^{*}$	&	$3.95$	&	 $\mathbf{3.99}$		\\ \hline
  
    \end{tabular}
\caption{Performance of different algorithms on an offline dataset. The results with $*$ denote the best performance among all baseline methods in each response dimension, and the data in last column is marked by bold font when our constrained-DDPG achieves the best performance. }
\label{tab:offlineV2}
\end{table*}

Table \ref{tab:offlineV2} presents the results of different algorithms in terms of eight scores. First note that A3C is outperformed by DDPG in most scores, which is as expected since A3C is an online learning algorithm that does not fit the offline setup here; this justifies our comparison focusing on DDPG-based constrained RL algorithms.  We can see that our  approach Constrained-DDPG performs the best among all algorithms: for the main goal, Constrained-DDPG achieves the highest overall rating $3.99$; for the auxiliary goal,  Constrained-DDPG also ranks highest for $5$ out of $7$  scores (service, cleanliness, value, rooms, location). The Pareto algorithm indeed learns a Pareto-optimal solution that achieves best performance on the check-in score, which however does not satisfy the setting here with the main goal to optimize the overall rating. The RCPO algorithm achieves the same best  overall score as our approach, but they sacrifice much on the others, and in particular, the location score is even lower than that from the BC algorithm.

\subsection{Ablation Study}

\begin{figure*}
    \centering
    \includegraphics[width=14cm]{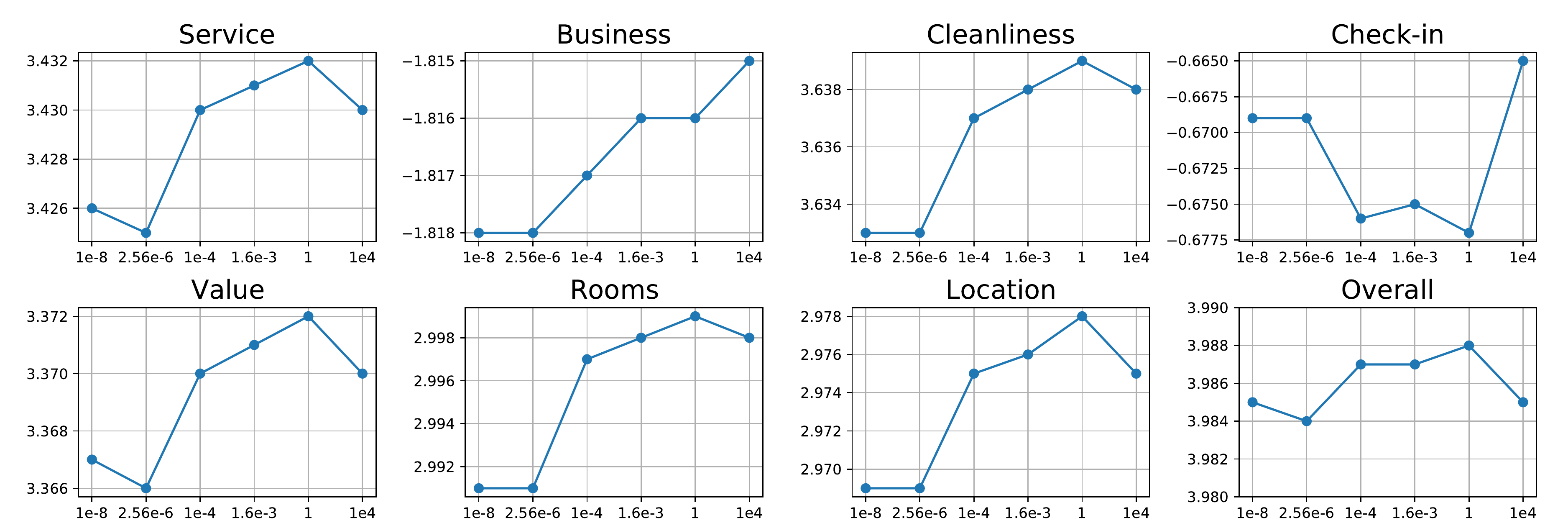}
    \caption{Effect of Lagrangian multiplier on model performance. The X-axis is the  of Lagrangian multiplier and the Y-axis is the score of each response.}
    \label{fig:offline-ablation}
\end{figure*}

The ablation study contains discussing the effect of the Lagrangian multiplier as well as the effect of the discount factor. Due to lack of space, the latter is attached in Appendix E. We investigate how the Lagrangian multiplier, which controls the strength of constraint, affects our model performance.  We vary $\lambda$ across $[1e-8, 2.56e-6, 1e-4, 1.6e-3, 1, 1e4]$ and present  performance of our constrained-DDPG in terms of all eight scores. Recall that larger $\lambda$ denotes stronger constraint that optimizes scores other than the overall one. Figure \ref{fig:offline-ablation} shows that with $\lambda$ increasing, our policy performance is also improved on most constraint scores (including service, business, cleanliness, value, rooms, and location), which is as expected since the learned policy becomes closer to the constraint policy that optimizes those scores. 

\section{Live Experiments}
The ultimate goal of recommender systems is to improve online user experience. To demonstrate the effectiveness of our algorithm, we test its real-world performance as well as other alternatives via A/B experiments. Algorithms are embodied in a candidate-ranking system used in production at a popular short video platform, that is, when a user arrives, these algorithms are expected to rank the candidate videos, and the system will recommend the top video to the user. 
We show that the proposed constrained actor-critic model is able to learn a policy that maximizes the main goal while also effectively balancing the auxiliary goal, and in particular, we set the main one as  maximizing the watch time and the auxiliary one as  improving the interactions between users and videos.

\subsection{Setup}
\paragraph{Evaluation metrics}
We use online metrics to evaluate policy performance. For the main goal, we look at the total amount of time user spend on the videos, referred to as \watchtime. For the auxiliary goal, users can interact with videos through multiple ways, such as sharing the video to friends, downloading it, or providing comments. Here, we focus on the three online metrics associated with the user-video interactions---the total number of \share, \download, \cmt~interactions.

\paragraph{MDP}
Following the formulation in Section \ref{sec:cmdp}, we present the constrained MDP in the context of short video recommendation. A trajectory starts when a user opens the app and ends when the user leaves. At time $t$, we have 
\begin{itemize}
    \item state $s_t$: a vector embedding of user current representation, for which we concatenate embeddings of  user historical interactions (encoded by recurrent neural networks) and  instantaneous context (such as device and location). 
    \item action $a_t$: a vector embedding of algorithm-predicted user preferences on different video topics, which determines the actual recommendation action--the video to be recommended---via a ranking function described below. 
    \item the ranking function: for each candidate video, this function calculates the dot product between the predicted user preference vector ($a_t$) and the video embedding (representing its topic and quality). The platform then recommends the video that achieves the largest score.
    \item reward $r_t=(l_t, i_t)$: after each recommendation, the system observes how long the user spent on the video, denoted as $l_t$, and  whether the user has interacted with the video (such as sharing/downloading/commenting on it), denoted as $i_t$.
    \item discount factor: we set $\gamma_l = 0.95$ for the time reward $l_t$ and $\gamma_i=0.0$ for interaction reward $i_t$ if not specified otherwise.\footnote{We find that $0.95$ is optimal for optimizing Watch time and $0$ is optimal for maximizing the interactions in live experiments.}
\end{itemize}

\paragraph{Compared algorithms} We choose A3C \cite{mnih2016asynchronous} as the base actor critic model, since algorithms compared are trained online in our live experiment setup, as opposed to the offline learning in Section \ref{sec:offline} that uses DDPG as the base actor critic model. 
Specifically, the action $a_t$ is sampled from a multivariate Gaussian distribution whose mean and variance are output of the actor model. We also complement our evaluation with  a supervised learning-to-rank (LTR) baseline, which is the default model run on the platform. 
\begin{itemize}
    \item \textbf{A3C}: We define a combined reward $m_t=l_t + \lambda i_t$ and learn a standard A3C \cite{mnih2016asynchronous} policy to maximize cumulative $m_t$ with discount factor $0.95$.
    \item \textbf{RCPO-A3C }: We separately learn two critic models  $V_l, V_i$ to evaluate cumulative time reward $l_t$ (with $\gamma_l=0.95$) and instant interaction reward $i_t$ (with $\gamma_i=0$). Then when optimizing the actor, we use advantage as a linear combination of  advantages calculated from two critic models respectively: $A_t = A_{l,t} + \lambda A_{i,t}$, where $A_{l,t} = l_t + \gamma_l V_l(s_{t+1}) - V_l(s_t)$, $A_{i,t} = i_t + \gamma_i V_i(s_{t+1}) - V_i(s_t)$, and $\lambda$ can be viewed as the Lagrangian multiplier in \cite{tessler2018reward}~\footnote{Different from \cite{tessler2018reward}, we here use a fixed Lagrangian multiplier since we found that in practice it is hard to specify the constraint level that is required to update Lagrangian multiplier in \cite{tessler2018reward}}.
    \item \textbf{Two-Stage constrained A3C (Ours)}: Following Algorithm \ref{alg}, we first learn a policy $\pi_i$ to optimize the auxiliary goal. 
    Then we learn a policy $\pi_d$ to optimize the main goal with the constraint that $\pi_d$ is in the neighborhood of $\pi_i$.
        \begin{itemize}
            \item \textbf{Interaction}: At the first stage, we  learn a A3C policy $\pi_i$ to maximize instant interaction reward $i_t$, with critic update following \eqref{eq:nonmajor-critic} and actor update following \eqref{eq:nonmajor-actor}.
            \item \textbf{Constrained} At the second stage, we learn a constrained A3C policy $\pi_d$ which  maximizes the cumulative time reward $d_t$ in the neighborhood of policy $\pi_i$, with critic update following \eqref{eq:major-critic} and  actor update following \eqref{eq:major-actor}. 
        \end{itemize}
        
    \item \textbf{LTR (Baseline)}: The learning-to-rank model\cite{liu2009learning} that takes user state embedding and video embedding as input and fits the sum of responses.
\end{itemize}


\paragraph{Experimental details} To test different algorithms, we randomly split users on the platform into five buckets with splitting ratio being $80\%, 5\%, 5\%, 5\%, 5\%$. The first bucket runs the baseline LTR model, and the remaining buckets run models A3C, RCPO-A3C, Interaction-A3C, and Constrained-A3C respectively. Models are pre-trained online for a couple of days and then are fixed to concurrently test  performance within one day.

\begin{table}[]
\label{table:result}
    \centering
    \begin{tabular}{p{2.3cm}|c|c|c|c}
    \toprule
        Algorithm &  \watchtime & \share &   \download & \cmt \\
        \midrule
        A3C & $+0.309\%$ & $ -0.707\%$ &  $0.153\%$ & $-1.313\%$ \\
        RCPO-A3C  & $+0.283\%$ & $-1.075\%$ &  $-0.519\%$ & $-0.773\%$ \\
        Interaction & $+0.117\%$ & $+5.008\%$ & $+1.952\%$ & $-0.101\%$ \\
        Constrained & $+0.336\%$ & $+3.324\%$ & $+1.785\%$ & $-0.618\%$ \\
         \bottomrule
    \end{tabular}
    \caption{Performance of different algorithms relative to a supervised LTR baseline in a live experiment.}
    \label{tab:live}
\end{table}

\subsection{Results}


Table \ref{tab:live} shows the results of algorithm comparison regarding metrics \watchtime, \share, \download, and \cmt. As we can see, both  A3C  with combined reward and RCPO-A3C  with combined advantage learn to improve the \watchtime as compared to the base model; but interaction-signal is too sparse with respect to \watchtime, such that when combining these two responses together--in the form of either reward or advantage--both models cannot effectively balance the interaction well. Performance of the Interaction model is as expected: with signal from only the interaction reward, the model learns to improve the interaction-related metrics (\share, \download, \cmt); such interactions between users and videos also improve the user watch time, since more interesting videos with high potential of invoking interactions are recommended, which optimizes user whole experience. Finally, our model achieves the best performance: as compared to  A3C  and  RCPO-A3C, it has slightly better \watchtime and  does much better on interaction metrics, thanks to the effective regularization during training  that it should not be too far from the Interaction-A3C policy. 
 
 \begin{figure}
    \centering
    \includegraphics[width=10cm]{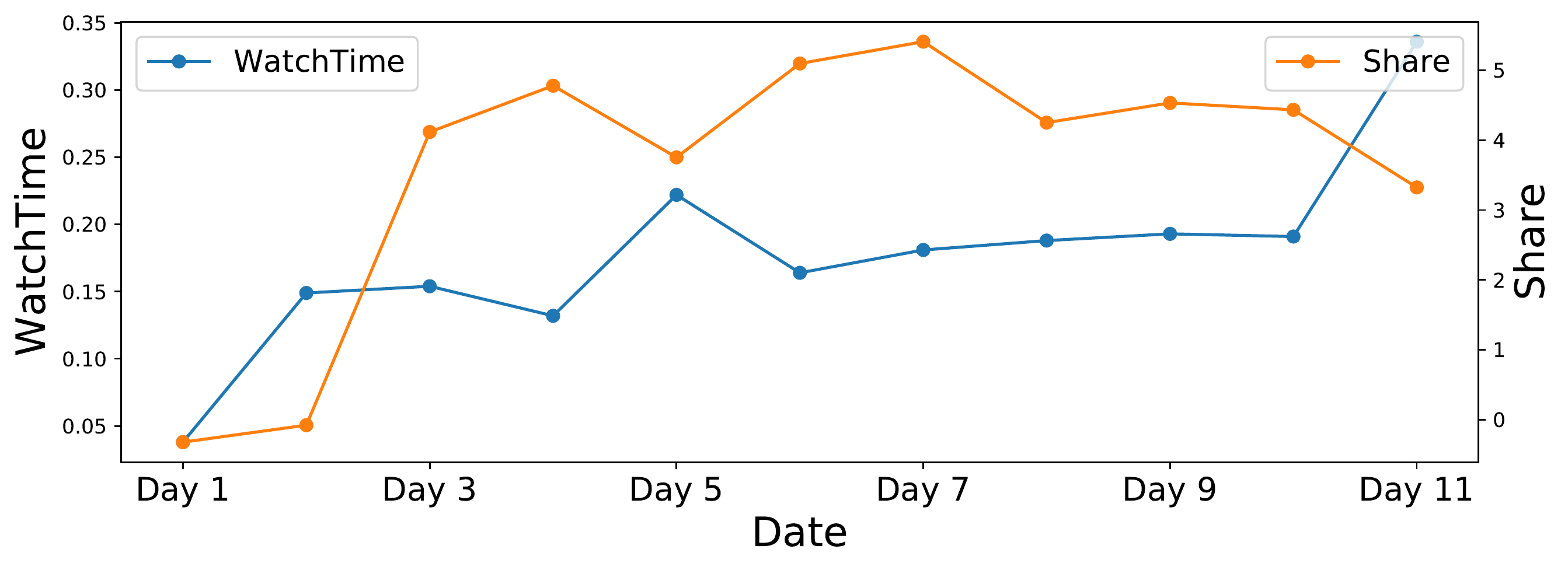}
    \caption{Learning curve of Contrained-Time-A3C. The blue and orange lines show its live performance on \watchtime~and \share~as compared to a supervised LTR baseline.}
    \label{fig:live_learning}
\end{figure}

To understand how our Constrained-A3C model learns to balance the main and auxiliary goal, Figure \ref{fig:live_learning}  plots its online performance--relative to the LTR baseline--during the learning phase on  the two live metrics: \watchtime  and \share. As shown, The model quickly learns to improve the \share metric by being restricted to the neighborhood of Interaction-A3C policy, demonstrating the effectiveness of our soft constraint. Then gradually, the model learns to improve \watchtime~over time. 

\section{Conclusion}

In this paper, we propose a constrained RL-based policy learning approach that optimizes the main goal as well as balancing the others for short video platforms. Our approach consists of two learning stages and is developed based on the actor-critic framework. At stage one, for each auxiliary response, we learn a policy to optimize its cumulative reward respectively.  At stage two, we learn the major policy that  optimizes the cumulative main response in the neighborhood of the policies learned at stage one for other responses. We demonstrate the advantages of our approach over existing alternatives via extensive offline simulations as well as live experiments. We show that our approach effectively achieves to optimize the main goal with a good trade-off for the auxiliaries. 


\bibliographystyle{ACM-Reference-Format}
\bibliography{reference}


\appendix
\section{Evaluation of Default Policy on A Popular Short Video Platform}
\label{appendix:offline-critic}
We showcase the advantage of separate evaluation for each response over a single evaluation of summed response. Specifically, we consider two types of responses from each video view: \watchtime~and interaction-signal (which is an indicator function of whether \share/\cmt/\download~happens during the  view).
\begin{itemize}
    \item For the single evaluation, we learn a value model $V_{single}$ with reward as a summation of \watchtime~and interaction-signal.
    \item For the separate evaluation, we learn two value models  $V_w$ and $V_i$ with reward as \watchtime~and interaction-signal respectively. Define the value  of separate evaluation as $V_{separate}=V_w + V_i$~\footnote{We focus on the summation of estimated values from $V_w$ and $V_i$, since it will be later used to guide policy learning}.
\end{itemize}
For fair comparison, we share the same discount factor $0.95$ for all value models and train them on the same data collected from a popular short video platform for one day. To evaluate the extend to which the value model learns the information of \watchtime~and interaction-signal, we compute the correlation between model values $V_{single}$ and $V_{separate}$ with the Monte Carlo value of the corresponding response. As compared to $V_{single}$, $V_{separate}$ is more correlated with \watchtime~and interaction-signal by $0.191\%$ and $0.143\%$ respectively, demonstrating that the separate evaluation better learns different reward responses.

\section{Derivation of the Constrained Actor-Critic Algorithm}
\label{appendix:derivation}
The Lagrangian of the constrained optimization \eqref{eq:awac} is 
\begin{equation}
\label{eq:awac_lagrangian}
     \mathcal{L}(\pi, \lambda_2,\dots, \lambda_m) = -E_{\pi}[ A_{1}^{(k)}] + \sum_{i=2}^m\lambda_i\big(
    D_{KL}(\pi|| \pi_{\theta_i}) - \epsilon_i
    \big),
\end{equation}
with $ \lambda_i \geq 0, i=2,\dots, m$ and $\int_a \pi(a|s)=1$. Compute the gradient of $ \mathcal{L}(\pi, \lambda_2,\dots, \lambda_m)$ with respect to $\pi$, we have
\begin{equation}
    \frac{\partial \mathcal{L}}{\partial \pi} = -A_1^{(k)} + \sum_{i=2}^m \lambda_i\big( 1 + \log(\pi(a|s)) - \log(\pi_{\theta_i}(a|s)) \big).
\end{equation}
Setting $ \frac{\partial \mathcal{L}}{\partial \pi}=0$, we have the solution $\pi^*$ satisfies
\begin{equation}
    \pi^*(a|s) = \frac{1}{Z(s)} \prod_{i=2}^m \big(\pi_{\theta_i}(a|s)\big)^{\frac{\lambda_i}{\sum_{j=2}^m \lambda_j}}\exp\bigg(\frac{A_{1}^{(k)}}{\sum_{j=2}^m \lambda_j} \bigg), 
\end{equation}
where $Z(s)$ is the partition function to make sure that $\int_a \pi(a|s)=1$.

To solve parameter $\theta_1^{(k+1)}$ of the main policy with data collected by $\pi_{\theta_1^{(k)}}$, we minimize the KL divergence between $\pi^*$ and $\pi_{\theta_1^{(k+1)}}$:
\begin{equation}
\begin{split}
     &\theta_1^{(k+1)}\leftarrow \arg\min_\theta E_{\pi_{\theta_1^{(k)}}}[D_{KL}(\pi^*(a|s)||\pi_\theta(a|s))]\\
     = & \arg\min_\theta E_{\pi_{\theta_1^{(k)}}} [E_{\pi^*}[-\log\pi_\theta(\cdot|s)]]\\
     = & \arg\max_\theta E_{\pi_{\theta_1^{(k)}}} [\bE_{\pi_{\theta_1^{(k)}}}[\frac{\pi^*(a|s)}{\pi_{\theta_1^{(k)}}(a|s)}\log\pi_\theta(\cdot|s)]]\\
     = &\arg\max_\theta E_{\pi_{\theta_1^{(k)}}}\Big[ 
     \prod_{i=2}^m \Big(\frac{\pi_{\theta_i}(a|s)}{\pi_{\theta_1^{(k)}}(a|s)}\Big)^{\frac{\lambda_i}{\sum_{j=2}^m \lambda_j}}\exp\bigg(\frac{A_{1}^{(k)}}{\sum_{j=2}^m \lambda_j} \bigg)
    \log \pi_\theta(a|s) \Big].
\end{split}
\end{equation}

\section{The Two-Stage Constrained Actor Critic Algorithm}

\begin{algorithm}[H]
\caption{Two-Stage Constrained Actor Critic}
\label{alg}
\textbf{Stage One:} For each auxiliary response $i=2,\dots,m$, learn a policy to optimize the response $i$, with  $\pi_{\theta_i}$ denoting actor and $V_{\phi_i}$ for critic.

While not converged, at iteration $k$:
\begin{small}
\begin{equation*}
    \begin{split}
       \phi_{i}^{(k+1)} \leftarrow  & \arg\min_{\phi}
    E_{\pi_{\theta_i^{(k)}}}\Big[
    \big(r_{i}(s, a) + \gamma_{i}V_{\phi_i^{(k)}}(s') -V_{\phi}(s)  \big)^2
    \Big],\\
        \theta_{i}^{(k+1)} \leftarrow & \arg\max_{\theta}  E_{\pi_{\theta_i^{(k)}}}\Big[ A_{i}^{(k)}\log\big(\pi_\theta(a|s)\big)\Big].
    \end{split}
\end{equation*}
\end{small}

\textbf{Stage Two:} For the main response, learn a policy  to both optimize the main response  and restrict its domain close to the policies $\{\pi_{\theta_i}\}_{i=2}^m$ of auxiliary responses, with $\pi_{\theta_1}$ denoting actor and $V_{\phi_1}$ for critic.

While not converged, at iteration $k$:
\begin{small}
\begin{equation*}
    \begin{split}
    \phi_{1}^{(k+1)} \leftarrow  & \arg\min_{\phi}
    E_{\pi_{\theta_1^{(k)}}}\Big[
    \big(r_{1}(s, a) + \gamma_{1}V_{\phi_1^{(k)}}(s') -V_{\phi}(s)  \big)^2
    \Big],\\ 
    \theta_1^{(k+1)}  \leftarrow  &\arg\max_{\theta}E_{\pi_{\theta_1^{(k)}}}\Big[ 
     \prod_{i=2}^m \Big(\frac{\pi_{\theta_i}(a|s)}{\pi_{\theta_1^{(k)}}(a|s)}\Big)^{\frac{\lambda_i}{\sum_{j=2}^m \lambda_j}}\\
    & \qquad\quad\quad\quad \times\exp\bigg(\frac{A_{1}^{(k)}}{\sum_{j=2}^m \lambda_j} \bigg)
    \log \pi_\theta(a|s) \Big].
    \end{split}
\end{equation*}
\end{small}

\textbf{Output:} the constrained policy $\pi_1$.
\end{algorithm}

\section{Offline Learning}
We now discuss adapting our constrained actor critic framework to policy learning using offline datasets. 

\paragraph{Multi-Critic Learning} In contrast to online learning that critics evaluating different responses are trained with data under different actor policies, these critics now share the same offline dataset, which can be formulated into a multi-task learning problem. One can apply off-the-shelf multi-task algorithms  to learn the commonalities shared by different tasks and also the differences for the task-specific considerations. As an example, the shared-bottom model architecture proposed in \cite{caruana1997multitask} is efficient in leveraging sharing attributes and avoiding overfitting, but it may bring in optimization conflict from different tasks for the bottom parameters. One remedy is to learn several expert models and use task-specific gates to generate each model prediction, named as the multi-gate mixture-of-experts (MMOE) method proposed in \cite{ma2018modeling}.

\paragraph{Debiased Actor Learning} The main change when moving from the online actor learning to the  offline setup is the bias correction on the policy gradient. The actor is no longer updated on data collected by current policy but by another behavior policy $\pi_\beta$, which may result in a different data distribution  induced by the policy being updated.  To address the distribution mismatch when estimating the policy gradient, a common strategy is to apply bias-correction ratio via importance sampling \cite{precup2000eligibility,precup2001off}. Given a trajectory $\tau=(s_1,a_1, s_2, a_2, \dots)$, the bias-correction ratio on the policy gradient for policy $\pi_{\theta_i}$ is
\begin{equation}
\label{eq:unbiased-is}
    w(s_t, a_t) = \prod_{t'=1}^t \frac{\pi_{\theta_i}(s_{t'}|a_{t'})}{\pi_\beta(s_{t'}|a_{t'})},
\end{equation}
which gives an unbiased estimation, but the variance can be huge. Therefore, we suggest using a first-order approximation of \eqref{eq:unbiased-is}, and use the current action-selection ratio when optimizing the actors of auxiliary responses, 
\begin{equation}
    \begin{split}
        \theta_{i}^{(k+1)} \leftarrow& \arg\max_{\theta}  E_{\pi_{\theta_i^{(k)}}}[ A_{i}^{(k)}\log(\pi_\theta(a|s))]\\
       & \approx   \arg\max_{\theta}  E_{\pi_\beta}\bigg[\frac{\pi_{\theta_i^{(k)}}(a|s)}{ \pi_\beta(a|s)} A_{i}^{(k)}\log(\pi_\theta(a|s))\bigg].
    \end{split}
\end{equation}
Similarly, when updating the actor of the main response, we have
\begin{equation}
    \begin{split}
         \theta_{1}^{(k+1)}  \leftarrow  &\arg\max_\theta \bE_{ \pi_{\theta_1^{(k)}}}\Big[ 
     \prod_{i=2}^m \Big(\frac{\pi_{\theta_i}(a|s)}{\pi_{\theta_1^{(k)}}(a|s)}\Big)^{\frac{\lambda_i}{\sum_{j=2}^m \lambda_j}}\\
      &\qquad\qquad\times
      \exp\bigg(\frac{A_{1}^{(k)}}{\sum_{j=2}^m \lambda_j} \bigg)
    \log( \pi_\theta(a|s)) \Big].\\
     &\approx\arg\max_\theta \bE_{ \pi_\beta}\Big[ 
     \prod_{i=2}^m \Big(\frac{\pi_{\theta_i}(a|s)}{\pi_{\theta_\beta}(a|s)}\Big)^{\frac{\lambda_i}{\sum_{j=2}^m \lambda_j}}\\
     &\qquad\qquad\times
      \exp\bigg(\frac{A_{1}^{(k)}}{\sum_{j=2}^m \lambda_j} \bigg)
    \log( \pi_\theta(a|s)) \Big].
    \end{split}
\end{equation}

\section{Ablation Study}

\begin{figure*}
    \centering
    \includegraphics[width=15cm, height=8cm]{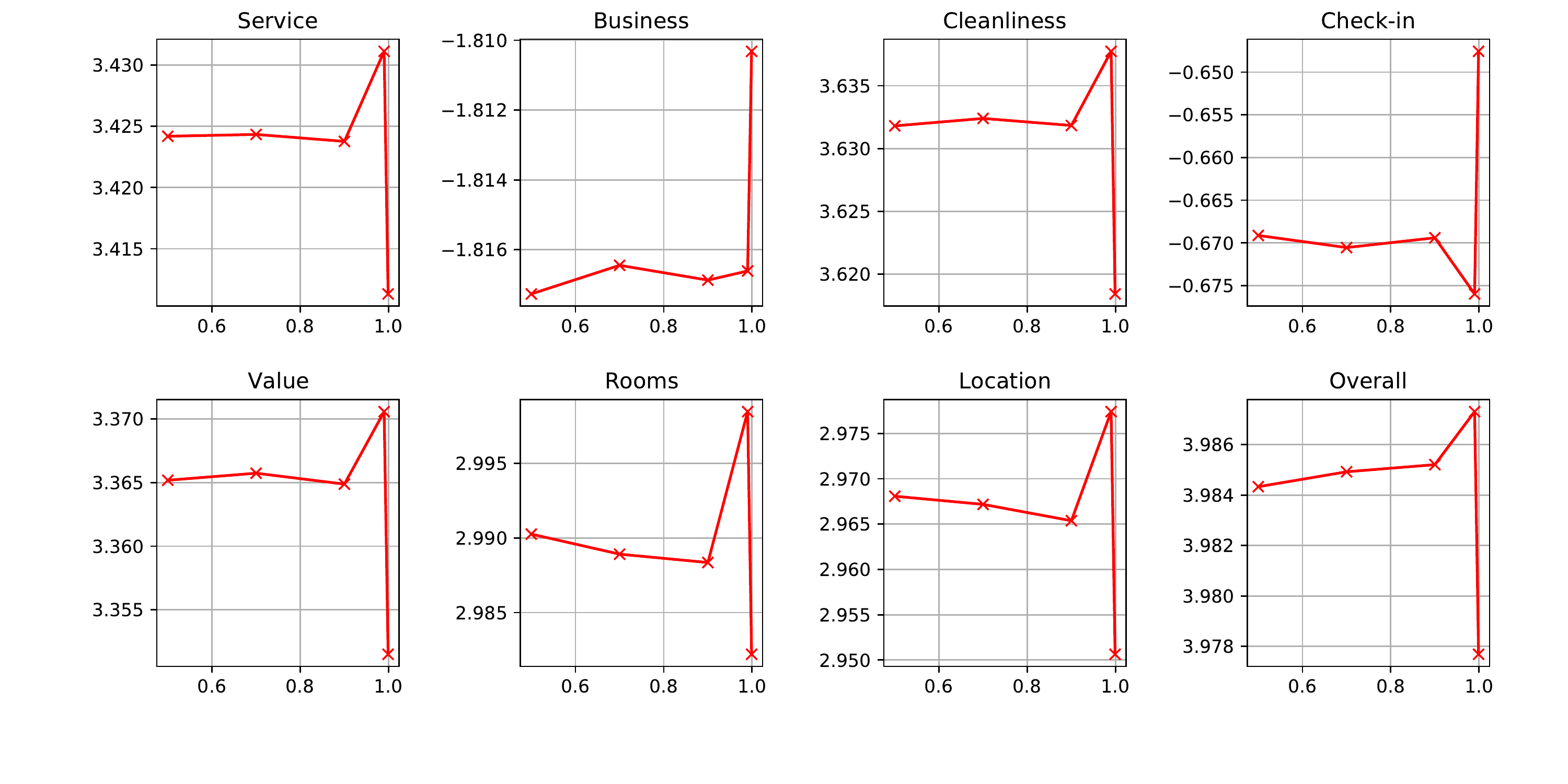}
    \caption{Effect of Discount Factor on model performance for all responses. The X-axis denotes the discount factor $\gamma$ and the Y-axis is the score of each response. }
    \label{fig:offline-gamma-full}
\end{figure*}

\paragraph{The Effect of the discount factor on the performance} As a critical hyper-parameter in RL methods, the discount factor $\gamma$ plays a vital role in model performance. To this end, we study the influence of $\gamma$ and range its value across $[0.5, 0.7, 0.9, 0.99, 0.999]$. 
We present our model performance on the main response (overall score) and other auxiliary tasks.
The results are shown in Figure \ref{fig:offline-gamma-full} . We can see that as $\gamma$ increases, our policy progressively delivers improved performance for the main response since the long term reward is further considered, while the performance on the auxiliary task first remains relatively stable and then peaks for $\gamma=0.99$. However, the policy performance may deteriorate when $\gamma$ is too large. In this dataset, We select $\gamma = 0.99$ as an appropriate choice.

\end{document}